\documentclass[twocolumn,10pt]{asme2ej}

\usepackage{graphicx} %% for loading jpg figures
\usepackage{hyperref}   % to set up hyperlinks

\usepackage{amsmath,amsfonts}
\usepackage[ruled,linesnumbered]{algorithm2e}
\usepackage{array}
\usepackage[caption=false,font=footnotesize,labelfont=rm,textfont=rm]{subfig}
\usepackage{textcomp}
\usepackage{stfloats}
\usepackage{url}
\usepackage{verbatim}
\usepackage{bm}
\usepackage{booktabs}
\usepackage{makecell}

\hypersetup{
	colorlinks=true,
	linkcolor=blue,
	citecolor=blue,
	urlcolor=blue,
}
\usepackage[square,numbers]{natbib}

\title{Multi-View Active Sensing for Human-Robot Interaction via Hierarchically Connected Tree}

\author{Yuanjiong Ying
    \affiliation{
	School of Mechanical Engineering\\
	Shanghai Jiao Tong University\\
	Shanghai, China, 200240\\
    Email: joyying0222@sjtu.edu.cn
    }	
}

\author{Xian Huang
	\affiliation{
		School of Mechanical Engineering\\
		Shanghai Jiao Tong University\\
		Shanghai, China, 200240\\
		Email: hx2020@sjtu.edu.cn
	}	
}

\author{Wei Dong
	\affiliation{
		School of Mechanical Engineering\\
		Shanghai Jiao Tong University\\
		Shanghai, China, 200240\\
		Email: dr.dongwei@sjtu.edu.cn
	}	
}

\begin{document}

\maketitle    

%%%%%%%%%%%%%%%%%%%%%%%%%%%%%%%%%%%%%%%%%%%%%%%%%%%%%%%%%%%%%%%%%%%%%%
\begin{abstract}
{\it Comprehensive perception of human beings is the prerequisite to ensure the safety of human-robot interaction. Currently, prevailing visual sensing approach typically involves a single static camera, resulting in a restricted and occluded field of view. In our work, we develop an active vision system using multiple cameras to dynamically capture multi-source RGB-D data. An integrated human sensing strategy based on a hierarchically connected tree structure is proposed to fuse localized visual information. Constituting the tree model are the nodes representing keypoints and the edges representing keyparts, which are consistently interconnected to preserve the structural constraints during multi-source fusion. Utilizing RGB-D data and HRNet, the 3D positions of keypoints are analytically estimated, and their presence is inferred through a sliding widow of confidence scores. Subsequently, the point clouds of reliable keyparts are extracted by drawing occlusion-resistant masks, enabling fine registration between data clouds and cylindrical model following the hierarchical order. Experimental results demonstrate that our method enhances keypart recognition recall from 69.20\% to 90.10\%, compared to employing a single static camera. Furthermore, in overcoming challenges related to localized and occluded perception, the robotic arm's obstacle avoidance capabilities are effectively improved.  
}
\end{abstract}

%%%%%%%%%%%%%%%%%%%%%%%%%%%%%%%%%%%%%%%%%%%%%%%%%%%%%%%%%%%%%%%%%%%%%%
\section{Introduction}

Collaborative robotic arms have been broadly deployed to assist humans  in the process of industrial production\cite{ref1}. When accomplishing manufacturing tasks, intensive interactions between humans and collaborative robots are virtually inevitable\cite{ref2, ref3}. To ensure the safety of human-robot interaction (HRI), the robot requires the ability to avoid dynamic human obstacles. Due to the extensiveness of task space and the sparsity of obstacles, mobile robots and aerial robots usually treat human as a bounding volume\cite{ref4,ref5}, without considering the detailed postures. However, since the workspace of robotic arm is restricted and immovable, a specific part of the human body can disrupt its task execution. Therefore, robotic arms should  perceive both the position and posture of human obstacles. To guarantee the safe and uninterrupted progress of the manufacturing process, comprehensive human sensing becomes evident.

Currently, a single stationary vision sensor is mostly employed for human sensing\cite{ref6}, such as monocular RGB, stereo and RGB-D cameras. Among them, RGB-D cameras, which capture both color and depth images simultaneously, have gained widespread use due to their rich information, compact design and cost-effectiveness\cite{ref7}. Unfortunately, limited field of view (FOV) is the inherent constraint of vision sensors. Pre-deployed fixed on or external to the robotic arm, the camera’s FOV is inherently localized and susceptible to occlusion. Not all body parts are visible in the images, and they may be self-occluded or occluded by robots or objects. This constrained perception ability impairs the recognition and avoidance of human obstacles.

To achieve comprehensive environmental perception, there are two commonly used approaches. One is to utilize multiple cameras\cite{ref8,ref15}, with the combined FOVs of these cameras exceeding that of a single camera. The other approach is active vision\cite{ref9,ref10}, wherein the camera is rotated or translated to capture information from more crucial regions. Consider a multi-camera active vision system (MCAV) that involves multiple manipulable cameras, its perception field has been expanded compared to a single fixed camera. The number and configuration of cameras typically depend on the application scenario of the robotic arm\cite{ref11,ref12}. MCAV not only enables the recognition of humans in different directions, but also broadens the solution space for robotic arm trajectory planning. It can dynamically manage camera viewpoints to focus on the crucial regions\cite{ref28}, such as human arms that may interfere with the operation of the robotic arm.

The RGB-D data obtained from multiple dynamic sources cannot be directly applied for human representation. Comprehensive human perception can only be achieved by extracting implicit semantics or establishing a unified model, enabling the effective integration of multi-source visual information. Existing works primarily fall into two directions: learning-based\cite{ref13} and model-based\cite{ref14} approaches. Multi-view learning-based methods automatically extract implicit human patterns from large datasets, enabling end-to-end human inference. Nevertheless, the lack of consideration for the dynamic and limited characteristics of the cameras' FOV makes the current 3D datasets unsuitable for our scenario. Multi-view model-based approaches involve the explicit modeling of human appearance, structure or specific motion. Using a priori knowledge of human anatomy, these methods provide precise and real-time alignment between models and images. Unlike learning-based detectors that offer an informed guess for occluded or indistinct parts, model-based estimation results are susceptible to occlusion and indistinctness. The aforementioned methods do not consider self-occlusion from the human body or external occlusion caused by the robotic arm, rendering them unsuitable for applications in HRI scenarios. Additionally, these approaches are ineffective in addressing dynamic and restricted FOV, impeding the efficient integration of the rich visual information captured by MCAV. Consequently, the design of a multi-view active sensing approach for human-robot interaction becomes imperative.

In our research, we propose a multi-view active sensing system based on a hierarchically connected tree structure. Firstly, an multi-view active vision mechanism is implemented by equipping cameras with additional degrees of rotation. Utilizing the captured multi-source RGB-D data, the presence of the keypoints are determined based on HRNet, and the model tree of human body is established. The inference of keypoints is then analytically elevated from 2D to 3D using depth image slices, providing prior pose information of existing keyparts. After extracting the point clouds of keyparts with a occlusion-resilient mask, ICP registration can be performed following the hierarchical order while the anatomical constraints are followed. The primary contribution of this paper are as follows:

\begin{enumerate}
	\item{A multi-camera active vision system that captures RGB-D data from multiple crucial regions;}
	\item{A hierarchically connected tree for integrating visual information from multi-view dynamic sources;}
	\item{An information extraction approach that is resilient to the occluded and localized FOV in HRI.}
\end{enumerate}

The remainder of this paper is structured as follows: section~\ref{section2} introduces the related work. In section~\ref{section3}, we present the preliminary modeling, along with the introduction of two foundational methods. Section~\ref{section4} introduce a integrated human sensing procedure based on a hierarchically connected tree structure. Simultaneously, we propose an analytical estimation of keyparts and keypoints that is resilient to occlusion and limited perception. Section~\ref{section5} encompasses the experimental setups, results and analysis. Finally, section~\ref{section6} provides the conclusion of this paper.

%%%%%%%%%%%%%%%%%%%%%%%%%%%%%%%%%%%%%%%%%%%%%%%%%%%%%%%%%%%%%%%%%%%%%%
\section{Related Work}
\label{section2}

Vision sensors possess the advantages of non-contact, lightweight, rich information and high accuracy. The visual information significantly assists the robotic arm to perform a variety of vision-based tasks, e.g. object grasp, target recognition, quality inspection, as well as localization, mapping and obstacle avoidance. Compared to a single camera, employing multiple cameras or active vision system enables more comprehensive environmental perception. A multi-camera configuration captures more visual information due to its broader FOV. \cite{ref15.5} mounts three cameras with static viewing areas for multi-human 3D pose estimation. To avoid unexpected obstacles, \cite{ref8} uses three stationary Kinect sensors at different orientations to construct a danger map. And \cite{ref15} employs one eye-in-hand camera and one eye-to-hand camera for citrus harvesting. Active vision systems control the acquisition of visual information by manipulating the viewpoints of cameras actively. \cite{ref9,ref10} detect and track objects with a rotatable camera focusing on crucial area. \cite{ref11,ref12} propose multi-camera active-vision systems for mobile robot navigation and object tracking, respectively. MCAV enables the robots to actively focus on several significant viewpoints through a planning strategy\cite{ref16}.

In order to integrate the multi-view visual information for human sensing, state-of-the-art methodologies fall into two main categories: the learning-based and the model-based methods. Learning-based approaches involve machine learning, particularly deep neural networks, to automatically capture complex human patterns in large datasets. The performance of directly inferring 3D state from 2D image remains unsatisfactory due to the ill-posed and ill-conditioned nature of this problem\cite{ref17}. Existing works\cite{ref18,ref19,ref20} primarily estimate 3D pose over a multi-view set of 2D projections of human in each camera image. Taking advantage of the multi-view setting, the 3D pose is recovered from the 2D inference\cite{ref21,ref22}. Nevertheless, these methods have not addressed the dynamic variation in camera’s FOV and demonstrate a sensibility to ambiguity and occlusion.

In contract to learning-based methods, model-based methods rely on a priori knowledge of human characteristics, such as appearance, spatial relations and behavior. Employing skinned multi-person linear model (SMPL), \cite{ref23,ref24} and \cite{ref25,ref26} employ optimization-based and regression-based methods, respectively, to estimate human pose and shape parameters from multi-view images. Based on a unconstrained 3D human shape model, \cite{ref27} performs multi-view pose estimation through single-frame recovery, temporal integration and model adaption. However, this inherent "matching" characteristic impairs the performance of model-based methods in the presence of occlusion or local perspectives, which are common in robotic arm scenarios.

%%%%%%%%%%%%%%%%%%%%%%%%%%%%%%%%%%%%%%%%%%%%%%%%%%%%%%%%%%%%%%%%%%%%%%
\section{Preliminaries}
\label{section3}

\subsection{Human 3D Body Model}

\begin{figure}[t]
	\centering
	\subfloat[Keypoints extraction]{
		\includegraphics[width=0.35\columnwidth]{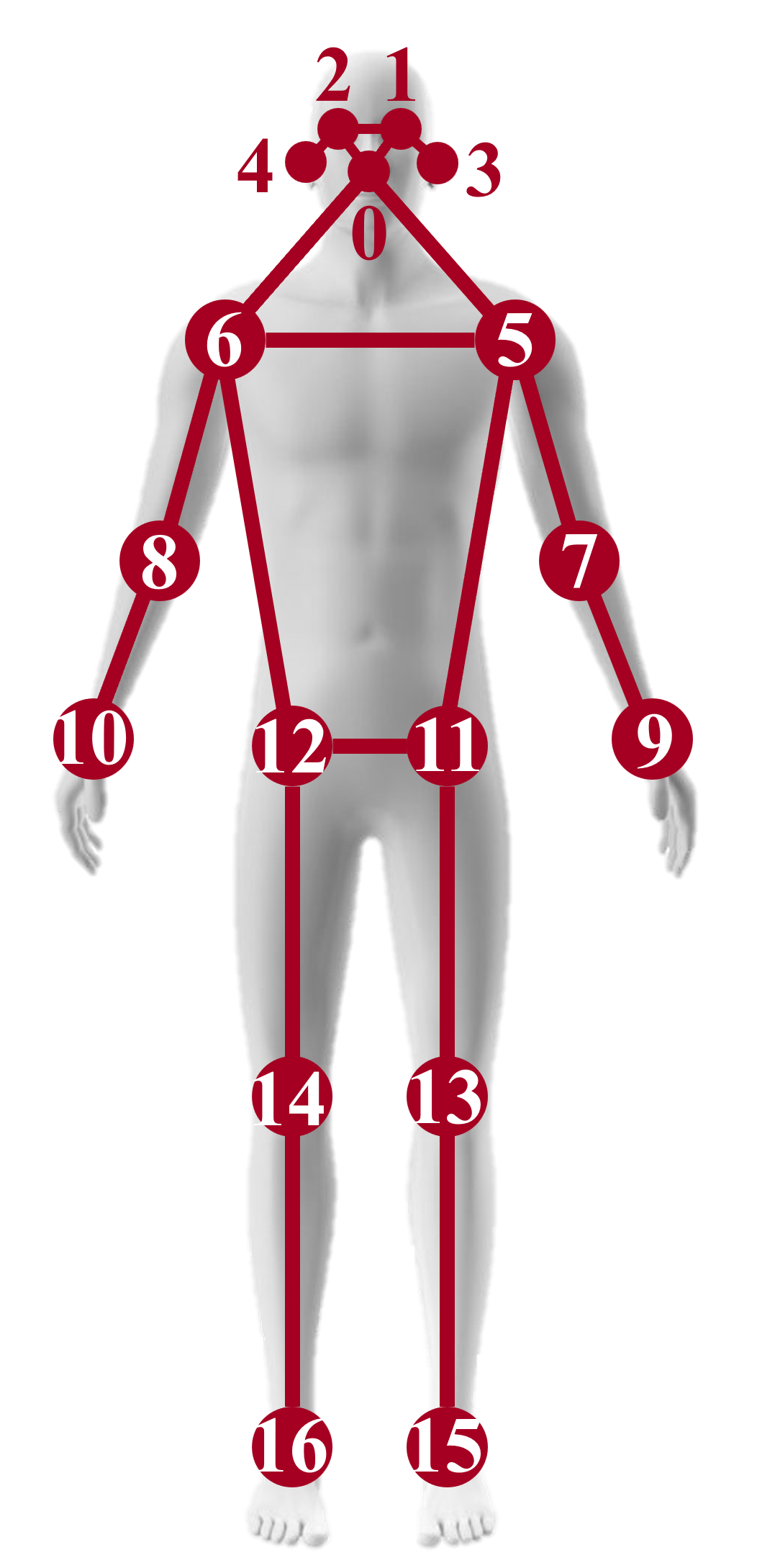}}
	\label{keypoints}
	\subfloat[Keyparts segmentation]{
		\includegraphics[width=0.55\columnwidth]{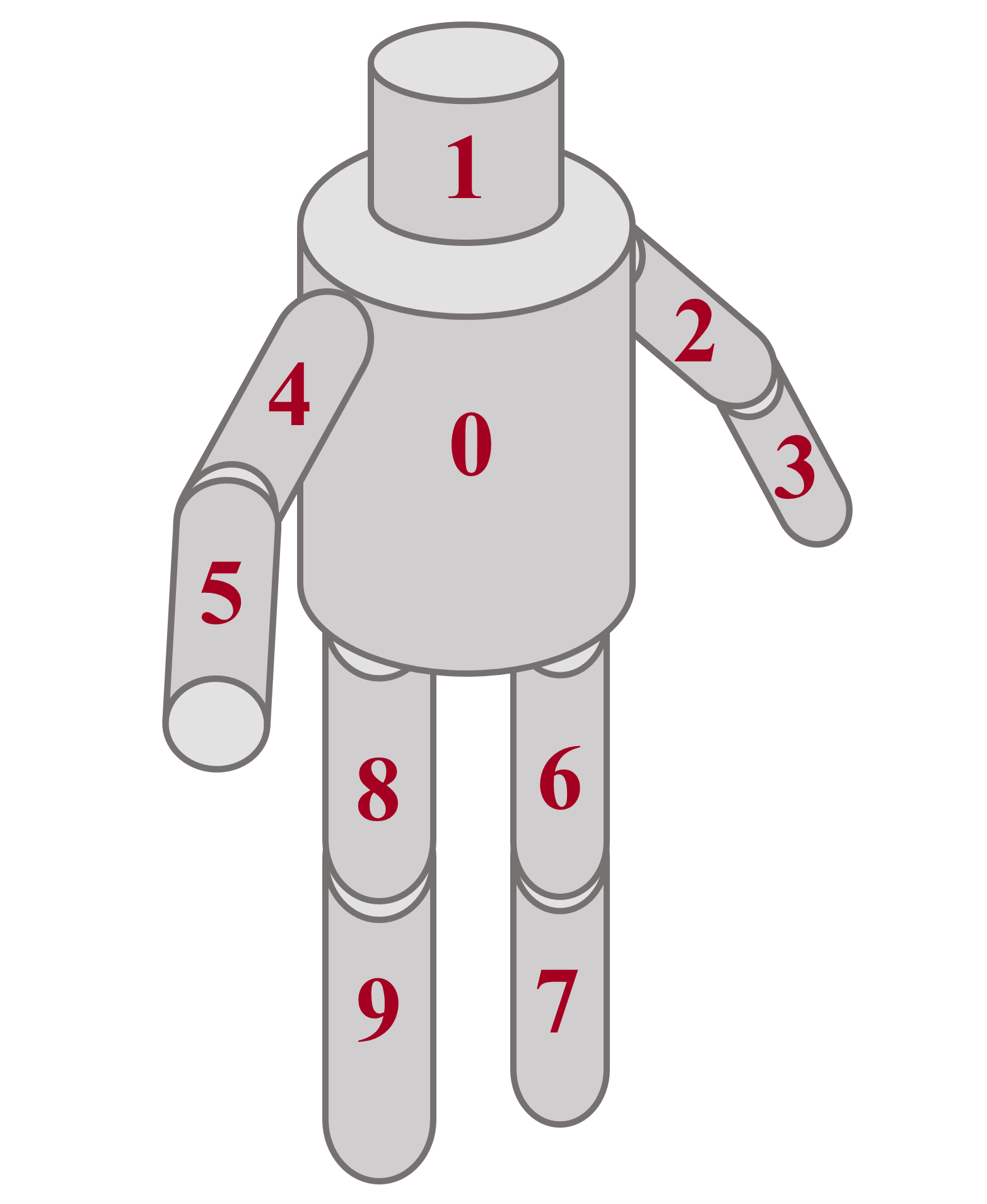}}
	\label{keyparts}
	\caption{Feature representation of human body model}
	\label{fig1}
\end{figure}

Due to the geometric characteristics of the human body, a cylindrical structural model is employed to represent human body, as illustrated in Fig.~\ref{fig1}. We extract 17 keypoints of the human body for preliminary estimation and state constraints, following the COCO dataset output format. And the entire body is segmented into 10 keyparts: a head, a torso, two upper arms, two lower arms, two upper legs, and two lower legs. Each keypart is modeled as cylinders with predefined radius and height.

Human body pose has high-dimensional and non-linear nature, where each limb's movement may affect other limb's movements. To reduce the Degrees of Freedom (DOF), we articulate connections between two adjacent keyparts through keypoint joints, disregarding the rotation of each keypart around its respective cylindrical axis. As depicted in Tab.~\ref{tab1}, both position and orientation of the torso in the world frame require estimation. As for other keyparts, their states depend on their parent part's state and their rotations around the joint. This human body model is structured as a hierarchical tree, with DOF reduced from 60 to 24.

\begin{table}[t]
	\caption{Human Body Modeling\label{tab1}}
	\centering
	\resizebox{1.0\columnwidth}{!}{
		\begin{tabular}{cccc}
			\toprule
			\textbf{Key parts} & \textbf{DOF} & \textbf{States}       & \textbf{Key points}     \\ \midrule
			Torso $\left\{0\right\}$ & 6 & $\left\{p^x,p^y,p^z,\theta^x,\theta^y,\theta^z\right\}$ & \{5,6,11,12\} \\
			Head $\left\{1\right\}$ & 2 & $\left\{\theta^x,\theta^y\right\}$ & \{0,1,2,3,4\}           \\
			Left arm $\left\{2,3\right\}$ & 4 & $\left\{\theta^x_u,\theta^y_u,\theta^x_l,\theta^y_l\right\}$ & \{5,7,9\}     \\
			Right arm $\left\{4,5\right\}$ & 4 & $\left\{\theta^x_u,\theta^y_u,\theta^x_l,\theta^y_l\right\}$ & \{6,8,10\}    \\
			Left leg $\left\{6,7\right\}$ & 4 & $\left\{\theta^x_u,\theta^y_u,\theta^x_l,\theta^y_l\right\}$ & \{11,13,15\} \\
			Right leg $\left\{8,9\right\}$ & 4 & $\left\{\theta^x_u,\theta^y_u,\theta^x_l,\theta^y_l\right\}$ & \{12,14,16\} \\ \bottomrule
		\end{tabular}
	}
\end{table}

\subsection{High-Resolution Network}

Prior to spatial human perception, we employ the High-Resolution Network (HRNet) in our work for 2D keypoint inference. HRNet falls into the top-down method, where human instances are initially detected, followed by the localization of keypoints. HRNet consists of multiple branches with varying resolutions, the multiscale fusions between the branches enable the generation of high-quality feature heatmaps. The maintenance of high-resolution information across the network layers renders HRNet suitable for our research.

Given a three-channel color image of size $\text{W} \times \text{H} \times 3$, HRNet infers heatmaps of each keypoint with size $\text{W}' \times \text{H}'$. The output set of HRNet is $\left\{\bm{H}_1,\bm{H}_2,\ldots,\bm{H}_\text{K}\right\}$, where $\text{K}$ represents the number of keypoints. Each heatmap $\bm{H}_k$ indicates the location confidence of the $k$th keypoint.

\subsection{Iterative Closest Point}

In order to fit the extracted point clouds of keyparts form to the cylindrical model, the Iterative Closest Point (ICP) algorithm is utilized for point cloud registration. To register the predefined model set $\bm{C}_\text{model}=\left\{\bm{x}_1,\cdots,\bm{x}_\text{N}\right\}$, and the captured data set $\bm{C}_\text{data}=\left\{\bm{y}_1,\cdots,\bm{y}_\text{N}\right\}$, the optimization problem for minimizing the registration error is formulated as:
\begin{equation}
	\boldsymbol{\chi} =\arg_{\boldsymbol{\chi}}\min \sum_{i=1}^{\text{N}} \| \bm{y}_i-\bm{T}\left(\bm{x}_i,\boldsymbol{\chi}\right)	\|^2
\end{equation}
where $\bm{x}_i \in \mathbb{R}^3$ and $\bm{y}_i \in \mathbb{R}^3$ denote the coordinates of the $i$-th model and data point, respectively. $\boldsymbol{\chi}$ represents the state variable of the human body, and $\bm{T}\left(\cdot\right)$ is the corresponding transformation function.

Given the initial state parameters, the alignment is refined based on establishing point correspondences between model and data clouds. By employing techniques such as Singular Value Decomposition (SVD), an analytical solution for the optimization can be obtained, and the parameters are subsequently updated. This iterative process continues until a convergence criterion is met, achieving precise spatial alignment.

\begin{figure*}[t]
	\centering
	%		\includesvg[width=1.0\textwidth]{Fig/frame.svg}
	\includegraphics[width=1.0\textwidth]{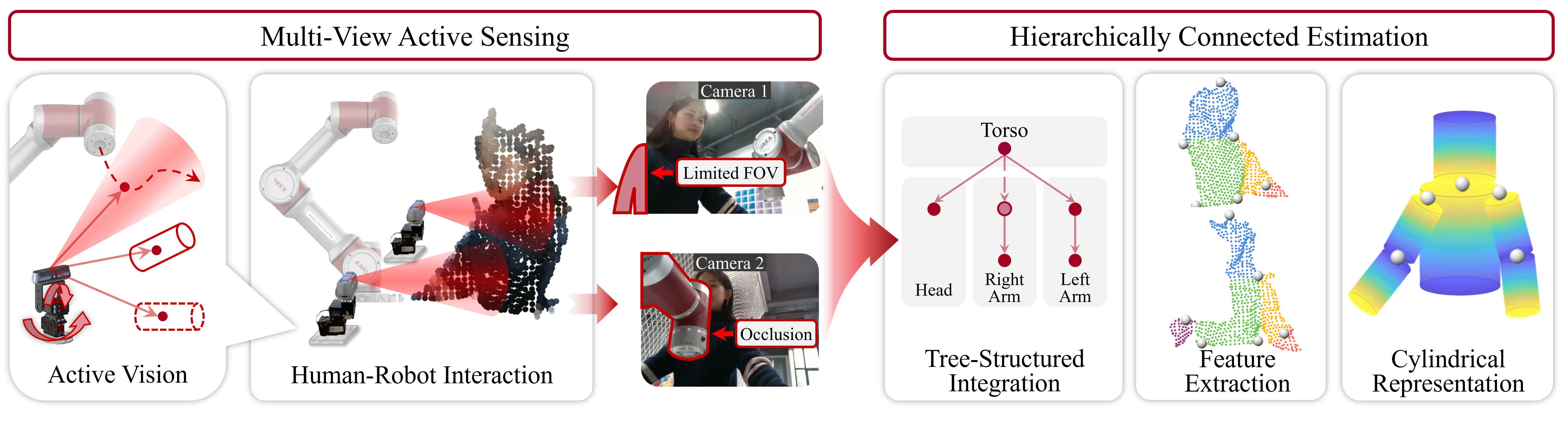}
	\caption{Overview of our multi-view active sensing system}        
	\label{frame}
\end{figure*}

%%%%%%%%%%%%%%%%%%%%%%%%%%%%%%%%%%%%%%%%%%%%%%%%%%%%%%%%%%%%%%%%%%%%%%
\section{Multi-view Active Human Sensing}
\label{section4}

Employing multi-camera active vision, our human sensing system for the HRI scenario is illustrated in Fig.~\ref{frame}. Firstly, MCAV manipulates each camera to focus on crucial regions, capturing a set of multi-view RGB-D data as input for the system. After determining the presence of each keypoint and keypart within each camera’s FOV, the connected directed tree of the human body can be constructed to integrate visual information. Subsequently, the 3D positions of keypoints are estimated, and the point clouds of keyparts are extracted. Finally, while preserving structural constraints, fine alignment between the cloud data and the cylindrical model is conducted following the hierarchical order.

\subsection{Hierarchically Connected Tree Model}

\begin{figure}[t]
	\centering
	\subfloat[Complete tree-structured model of human body]{
		\includegraphics[width=1.0\columnwidth]{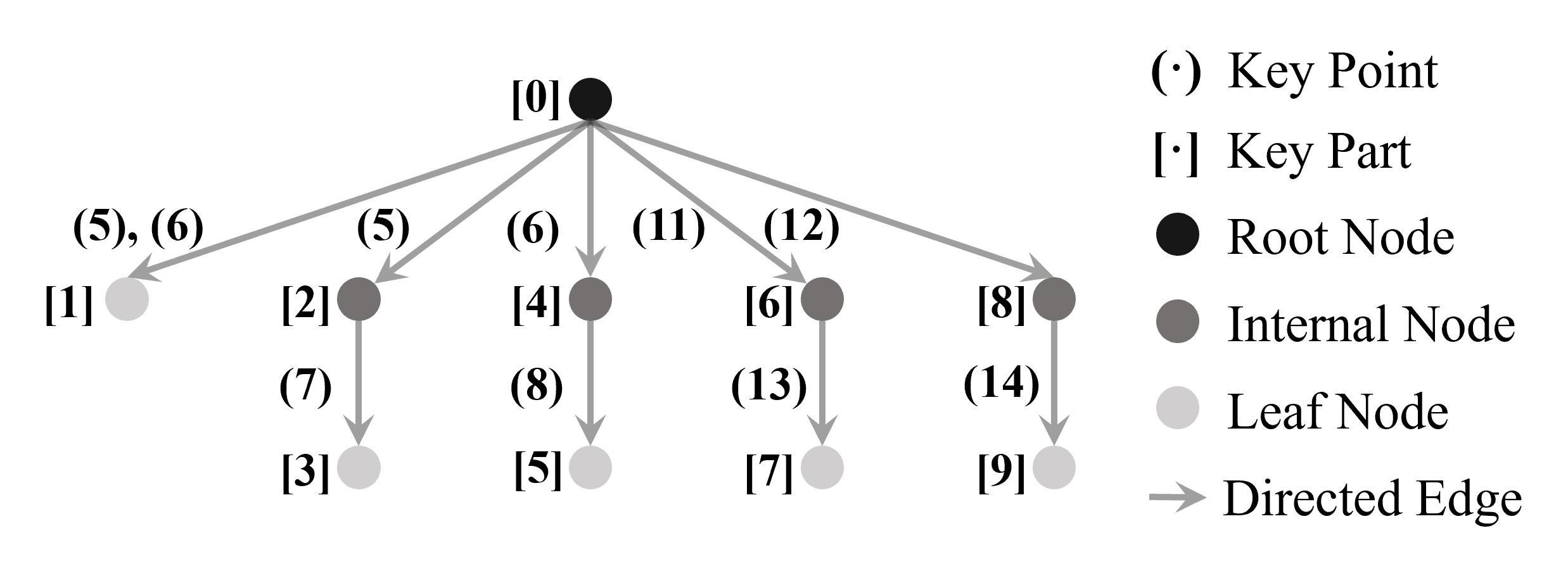}
		\label{fig21}}\\
	\subfloat[Connectivity maintenance of tree model with node supplementation]{
		\includegraphics[width=1.0\columnwidth]{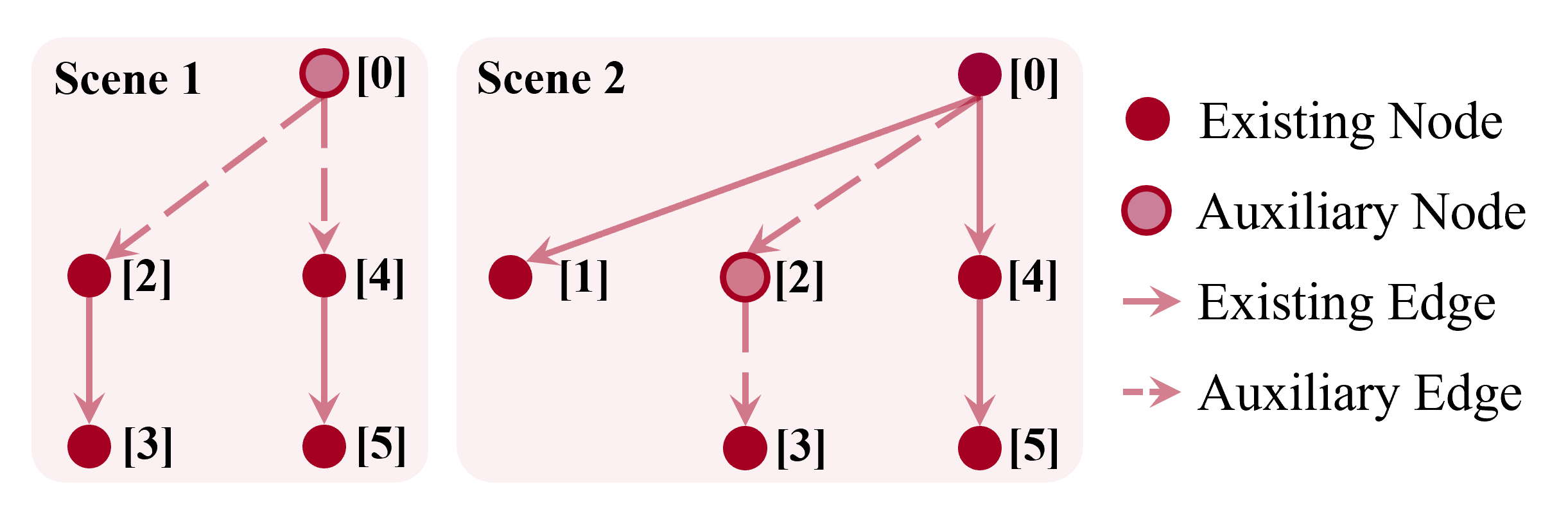}
		\label{fig22}}
	\caption{Hierarchically connected tree structure of human body}
	\label{fig2}
\end{figure}

In our research, we extract keypoints as features of the human body and partition the human body into keyparts. During human perception, estimating the states of each keypoint and keypart independently without considering anatomical structure is a straightforward approach. However, this may yield unsatisfactory results in practical applications, such as “disarticulated” human pose. Additionally, it hinders the effective fusion of visual data from each dynamic camera. Therefore, we structure the human body as a directed tree to maintain anatomical constraints, as well as integrate multi-source information. As depicted in Fig.~\ref{fig21}, each node of the tree structure represents a distinct keypart, with its presence determined by its observation status. Adjacent keyparts are interlinked through their corresponding keypoints, delineated by the directed edges. The nodes can be classified into root nodes, internal nodes, and leaf nodes, with the torso serves as the root node that links to the limbs and the head.

During HRI, the projection of the human operator on each camera plane may be partial and occluded, resulting in the nonexistence of specific nodes within the tree structure. Combining these subsets of tree model from each camera might compromise the overall structural connectivity. As illustrated in Fig.~\ref{fig22}, the absence of certain nodes leads to the fragmentation of the tree structure into multiple disconnected subtrees. Disconnectedness disrupts the intrinsic anatomical structure; hence, we introduce supplementary nodes to preserve connectivity. The supplementary node functions as either the root node or internal nodes. For the root node, its state can be approximated by the positions of the respective adjacent keypoints; whereas, for internal nodes, their states are determined by the respective keypoint positions with both parent and child nodes.

In order to estimate human body poses while maintaining anatomical constraints, we perform a hierarchical procedure following the direction of edges. The keypoints corresponding to each directed edge serve as a rigid constraint for joint connections, ensuring that the child keypart articulates with its parent keypart. Moreover, the progress of subsequent estimations rely on the existing estimate result. ICP algorithm between the model and the data clouds is employed for state estimation of each keypart. In the iterative registration process, the keyopints spatially localize the child keypart as the initial coarse state. And the alignment result updates the state of both the keyparts and their corresponding keypoints.

In summary, the connectivity-maintained hierarchical pose estimation procedure proposed in our research follows three primary principles: 
\begin{enumerate}
	\item{specific nonexistent nodes are supplemented to maintain the connectivity of the tree model;}
	\item{state estimation is performed following the hierarchical sequential order within the directed tree structure;}
	\item{available estimation results are utilized for the iterative initialization and the correction of anatomical constraints.}
\end{enumerate}

\subsection{Keypoint State Estimation}

The preliminary representation of human body relies on the state of keypoints, including their spatial positions and presence. The 2D locations and confidence scores of keypoints are inferred through HRNet from the color images, denoted as:
\begin{equation}
	\bm{p}_k = \underset{\substack{(u_k,v_k)\in \\ [1,\text{W}'] \times [1,\text{H}']}}{\arg\max} \bm{H}_k(u_k,v_k)
\end{equation}
\vspace{-2.5em}
\begin{equation}
	c_k = \underset{\substack{(u_k,v_k)\in \\ [1,\text{W}'] \times [1,\text{H}']}}{\max} \bm{H}_k(u_k,v_k)
\end{equation}
$\bm{p}_k \in \mathbb{R}^2$ represents the coordinates of the $k$th keypoint in the image plane, denoting the position of the heatmap peak. And $c_k \in \mathbb{R}$ represents the corresponding confidence score, signifying the peak value of the heatmap. The range of $c_k$ is constrained to $(0,1)$, and its magnitude effectively indicates whether the key point of the human body is occluded or outside the FOV. $k \in \left\{1,2,\ldots,\text{K}\right\}$, where $\text{K}$ is the number of keypoints. The inference result of $\text{K}$ keypoints can be represented as follows:
\begin{equation}
	\bm{p} = \left\{\bm{p}_1,\bm{p}_2,\ldots,\bm{p}_\text{K}\right\}, \quad \bm{c} = \left\{c_1,c_2,\ldots,c_\text{K}\right\}
\end{equation}
where $\bm{p} \in \mathbb{R}^{2 \times \text{K}}$, $\bm{c} \in \mathbb{R}^{1 \times \text{K}}$.

In the operational environment of robotic arms, the camera often captures the occluded or localized view. If a keypoint is occluded or located outside the FOV, its inference result becomes less reliable. Considering the confidence score, we propose a sliding-window-based formula for the presence determination of keypoints, expressed as:
\begin{equation}
	E_k=\frac{\text{sgn}\left(\sum_{m=0}^{\text{M}}\gamma^mc_{k,[t_{n-m}]}-\alpha\right)+1}{2}
\end{equation}
where $E_k\in\left\{0,1\right\}$, with 0 representing absence and 1 representing presence. $\text{sgn}(\cdot)$ is the signum function, $c_{k,[t_{n}]}$ denotes the confidence score of the $k$th keypoint at time $t_{n}$. A weighted sliding window of length $\text{M}$ is employed to mitigate inference errors at the current time. $\gamma \in \left(0,1\right)$ is the forgetting factor, and $\alpha \in \left(0,\text{M}\right)$ serves as the corresponding threshold limit. This formula determines the usability of the inferred positions for each keypoint, and the set of existing keypoints in the $i$th camera's FOV can be denoted as $\mathcal{K}_i=\left\{ k | k \in \left\{1,2,\ldots,\text{K}\right\},E_k = 1 \right\}$.

An RGB-D camera captures color and depth images simultaneously, with a one-to-one correspondence between the pixels in the images. In order to elevate the positions of keypoints from 2D to 3D in the camera frame, we use the depth image slices to estimate the depth data using the formula:
\begin{equation}
	d_k=\frac{1}{\text{N}_\text{r}} \sum_{\|\left(u_i,v_i\right)-\bm{p}_k\|<\text{r}} \bm{D}\left(u_i,v_i\right) \quad \text{for }k\in\mathcal{K}_i
\end{equation}
where $d_k \in \mathbb{R}$ represents the mean depth around the $k$th keypoint, for the depth of the inferred pixel may exhibit significant noise or absence of value. $\text{r}$ is the neighborhood radius, and $\text{N}_{\text{r}}$ is the number of the neighborhood points. Given the depth information and the pixel coordinates, the keypoint can be reprojected into the 3D space as follows:
\begin{equation}
	\bar{\bm{p}}_k^\mathcal{C} = d_k \mathbf{K}^{-1} \hat{\bm{p}}_k\quad \text{for } k\in\mathcal{K}_i
\end{equation}
where $\hat{(\cdot)}$ denotes adding a $1$ at the back of the vector, $\bar{(\cdot)}$ denotes the 3D coordinates reprojected from the 2D plane. Thus $\hat{\bm{p}}_k=\begin{bmatrix}u_k&v_k&1\end{bmatrix}^\text{T}$ is the homogeneous coordinates of $\bm{p}_k$, denoting the 2D projection of the $k$th key point in the image plane. $\bar{\bm{p}}_k^\mathcal{C} \in \mathbb{R}^3$ is the 3D position of the $k$th key point in the camera frame $\mathcal{C}$, and $\mathbf{K}=\begin{bmatrix}
	f_x & 0 & c_x \\
	0 & f_y & c_y \\
	0 & 0 & 1
\end{bmatrix}$ is the camera's internal parameters.

As multiple active cameras are employed in our research, we propose a general method for integrating key point information from the cameras as follows:
\begin{equation}
	\bar{\bm{p}}_k^* =\underset{\bar{\bm{p}}_k^*}{\arg\min} \sum_{k\in\mathcal{K}_i}c_{k,i} \|\bm{T}_{\mathcal{C}_i}^{\mathcal{W}}\hat{\bar{\bm{p}}}_k^{\mathcal{C}_i} -\hat{\bar{\bm{p}}}_k^{*}\|^2
\end{equation}
The equation above reveals that we employ an optimization approach to fuse the position data from multiple cameras, considering the confidence scores of keypoints as weights. And $c_{k,i}$ denotes the confidence score of the $k$th key point inferred by the $i$-th camera. $\bar{\bm{p}}_k^* \in \mathbb{R}^3$ represents the fused 3D position of the $k$th keypoint in the world frame $\mathcal{W}$, while $\bar{\bm{p}}_k^{\mathcal{C}_i}$ denotes the inferred 3D position of the $k$th keypoint in the respective camera frame $\mathcal{C}_i$. The homogeneous coordinates of $\bar{\bm{p}}_k^{\mathcal{C}_i}$ and $\bar{\bm{p}}_k^*$ are represented by  $\hat{\bar{\bm{p}}}_k^{\mathcal{C}_i}$ and $\hat{\bar{\bm{p}}}_k^{*}$, respectively. The transformation matrix from $\mathcal{C}_i$ to $\mathcal{W}$,  denoted as $\bm{T}_{\mathcal{C}_i}^{\mathcal{W}}$, is obtained through real-time knowledge of each camera's pose. If a keypoint presents in at least one camera's FOV, its 3D position can be estimated. Thus, the set of existing key points is denoted as $\mathcal{K}_\text{E} = \bigcup_{i=1}^{\text{N}_\text{c}} \mathcal{K}_i$, where $\text{N}_\text{c}$ is the number of the cameras. As this is an unconstrained quadratic optimization problem, an analytical solution is attainable and can be formulated as follows:
\begin{equation}
	\hat{\bar{\bm{p}}}_k^{*} = \frac{\sum_{i=1}^{\text{N}_k}c_k^i\bm{T}_{\mathcal{C}_i}^{\mathcal{W}}\hat{\bar{\bm{p}}}_k^{\mathcal{C}_i}}{\sum_{i=1}^{\text{N}_k}c_k^i}
	\quad \text{for }k \in \mathcal{K}_\text{E}
\end{equation}
where $\text{N}_k$ represents the count of existing $k$th keypoints. And the updated weight can be denoted as
\begin{equation}
	c^*_k = \frac{1-e^{-\text{N}_k}}{1+e^{-\text{N}_k}}\left(\frac{1}{\text{N}_k}\sum_{i=1}^{\text{N}_k}c_{k,i}\right) \quad \text{for }k \in \mathcal{K}_\text{E}
\end{equation}
where $c^*_k \in (0,1)$ denotes the confidence score of the fused keypoint, $\frac{1}{\text{N}_k}\sum_{i=1}^{\text{N}_k}c_{k,i} \in (0,1)$ is the average confidence score. $\frac{1-e^{-\text{N}_k}}{1+e^{-\text{N}_k}} \in (0,1)$ serves as the effectiveness factor, which increases with the growth of $\text{N}_k$, but does not increase unboundedly.

\subsection{Keypart Point Cloud Extraction}

In our research, the degenerative projections of the human body in the image plane are represented by ensembles of keyparts, with each keypart encompassing a set of keypoints. Therefore, based on the state estimation results of keypoints, we can conduct presence determination, point cloud extraction, and state initialization for each keypart.

During the manufacturing process, the dynamic cameras may focus on different portions of the environment. Consequently, a keypart that is entirely within the FOV of one camera might fall outside the FOV of another. Hence, the existence of each key part should be determined independently. We propose a formula with a structure similar to that of keypoints, represented as follows:
\begin{equation}
	E_j=\frac{\text{sgn}\!\left(\!\sum_{m=0}^{\text{M}}\gamma^mc_{\left(j\right),[t_{n-m}]}\!-\!\alpha\!\right)\!+\!1}{2}
\end{equation}
where $E_j \in \left\{0,1\right\}$ represents the existence value of the $j$th keypart. $j \in \left\{1,2,\ldots,\text{J}\right\}$, where $\text{J}$ is the number of keyparts. $\gamma \in \left(0,1\right)$ and $\alpha \in \left(0,\text{M}\right)$ are the corresponding forgetting factor and threshold limit, respectively. $c_{\left(j\right)} = \max\left\{c_k \mid k \in \mathcal{K}_{\left(j\right)} \right\}$ is the maximum confidence score among the keypoints compassed by the $j$th keypart, denoted by $\mathcal{K}_{\left(j\right)}$. The set of the existing keyparts can be denoted as $\mathcal{J}=\left\{ j | j \in \left\{1,2,\ldots,\text{J}\right\},E_j = 1 \right\}$.

The depth images from the RGB-D camera effectively capture the 3D spatial information of the surroundings. We have utilized this depth information to estimate and fuse the spatial positions of keypoints. To obtain the state of each human body part, we will make masks and apply them to the depth image. This procedure aims to extract the point cloud of the keyparts within the camera's FOV, thereby enabling a fine estimation of their positions and orientations.

\begin{figure}[t]
	\centering
	\subfloat[Keypoint projection]{
		\includegraphics[width=0.31\columnwidth]{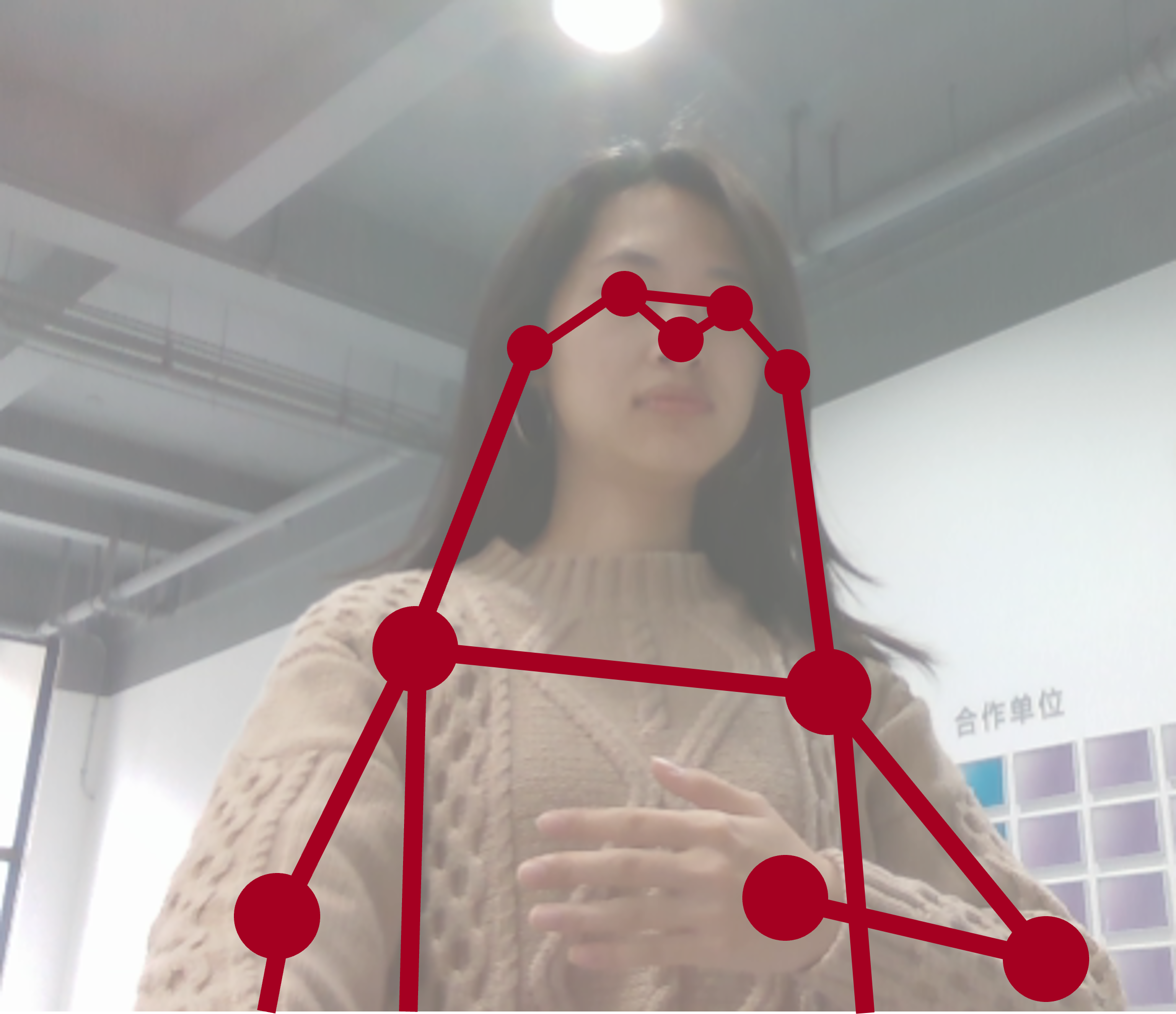}
		\label{mask1}}
	\subfloat[Mask generation]{
		\includegraphics[width=0.31\columnwidth]{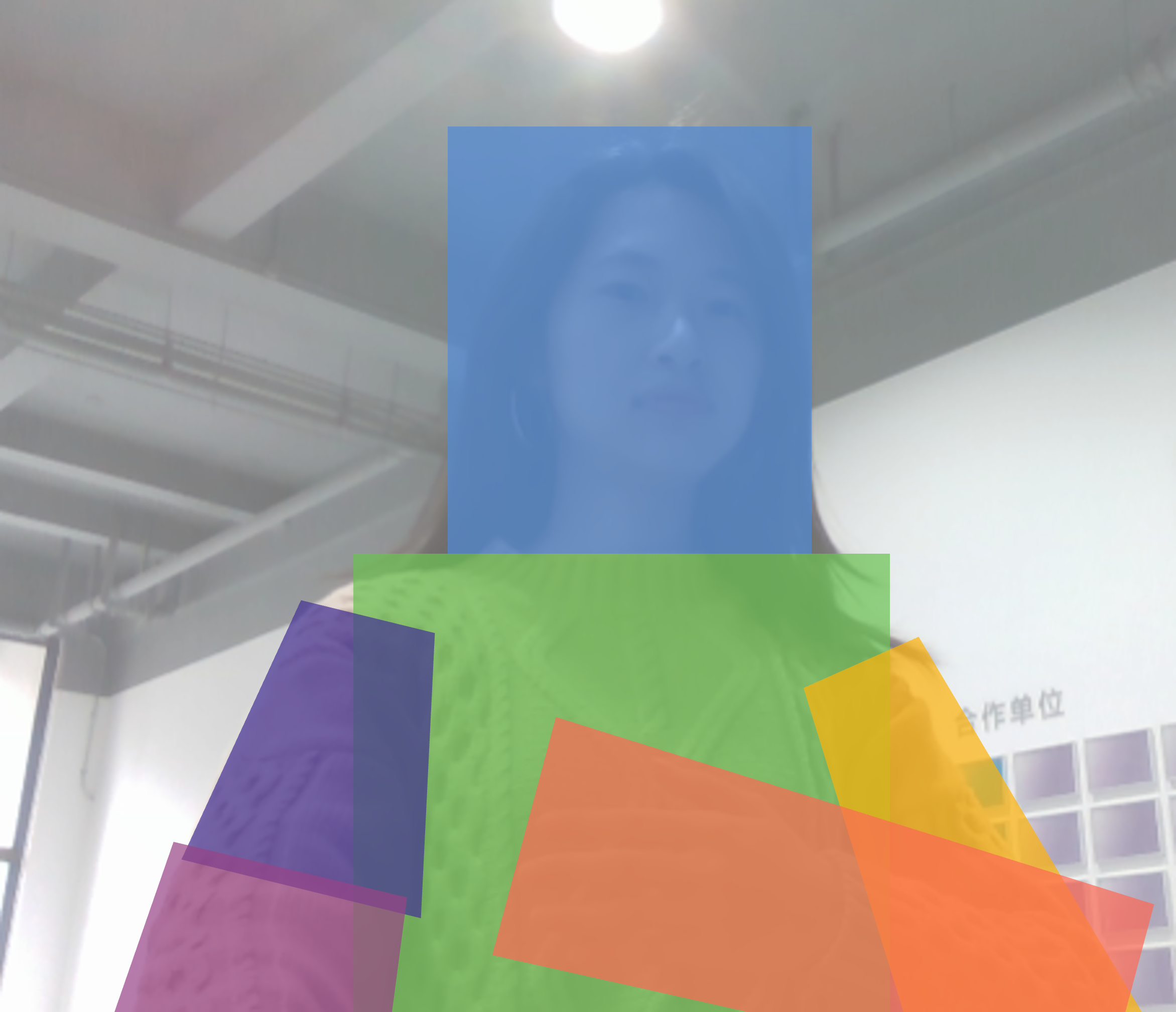}
		\label{mask2}}
	\subfloat[Cloud extraction]{
		\includegraphics[width=0.28\columnwidth]{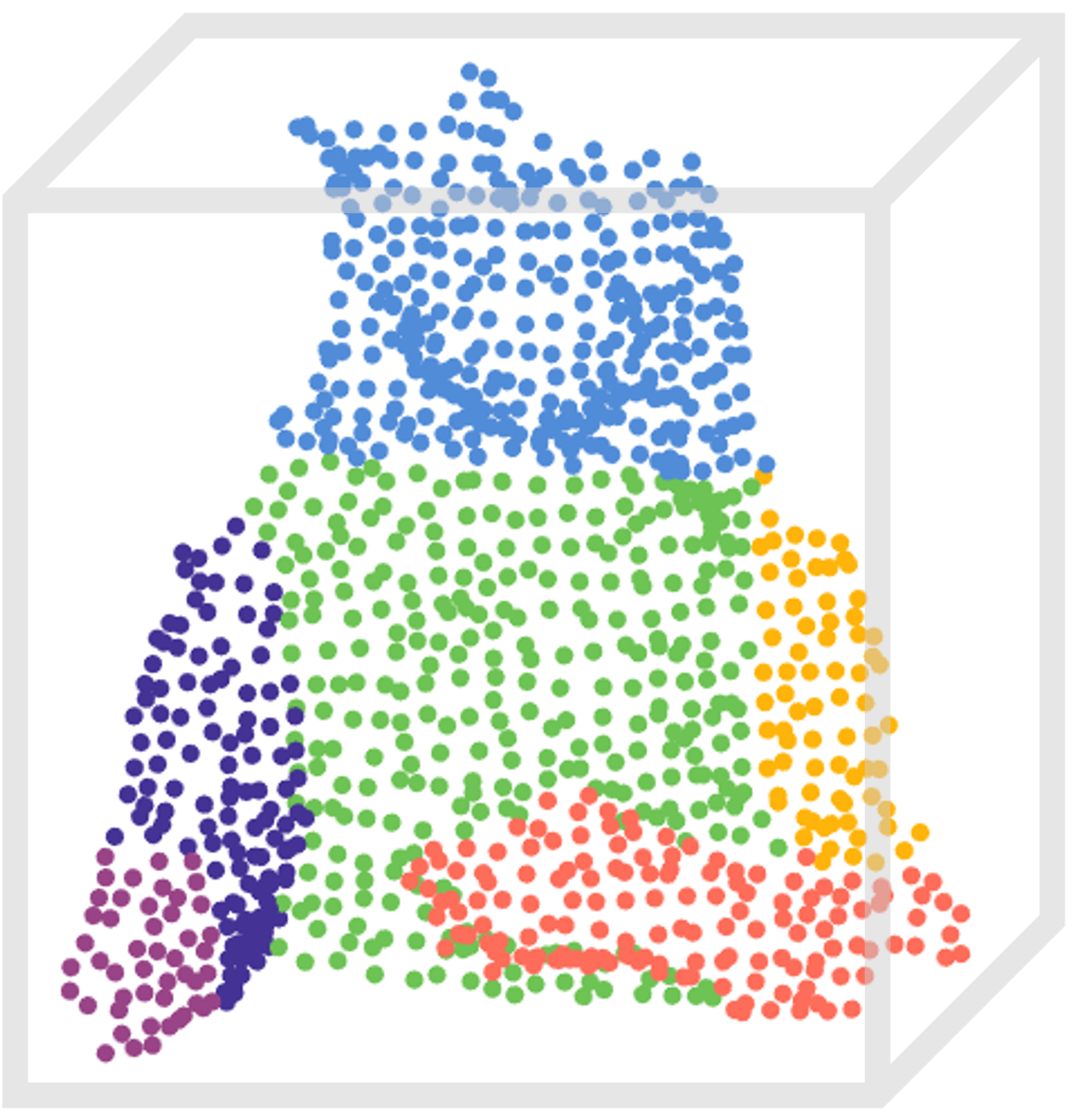}
		\label{mask3}}
	\caption{Point cloud extraction of existing keyparts}
	\label{mask123}
\end{figure}

To create masks for each camera, we localize the keypoints on the mask plane, including reliable ones and unreliable ones. Thus, for each $j \in \mathcal{J}$, we have
\begin{equation}
	\hat{\bm{p}}_{k,i}^* = 
	\begin{cases} 
		\left(d^*_{k,i}\right)^{-1}\mathbf{K}_i \bar{\bm{p}}_{k,i}^{*},
		& k \in \mathcal{K}_{\left(j\right)} \cap \mathcal{K}_\text{E}
		\\
		\hat{\bm{p}}_{k,i},
		& k \in \mathcal{K}_{\left(j\right)} \cap \left( \lnot \mathcal{K}_\text{E} \right)
	\end{cases}
\end{equation}
Since one existing keypoint implies the corresponding keypart's presence in the FOV, one existing keypart may have unreliable keypoints. The equation above indicates that the fused key points are projected onto the image plane, and the absent ones are replaced by the inference from the HRNet.

Based on the assumption that each keypart model is represented as a cylinder, its projection in the image plane can be approximately enveloped by an isosceles trapezoid. The shape and position of the isosceles trapezoid can be determined by four parameters: the midpoints and lengths of its upper and lower bases. The midpoints of the upper and lower bases can be inferred from the localization of keypoints. Meanwhile, the length of the base can be determined as follows:
\begin{equation}
	L_{j,b}=\frac{f}{d_{j,b}} \cdot R_j \quad \text{for }j \in \mathcal{J}, b \in \left\{\text{upper},\text{lower}\right\}
\end{equation}
where $L_{j,b}$ denotes the length of the base, $f$ is the focal length of the camera, $R_j$ is the predefined radius of the cylinder model part, and $d_{j,b}$ denotes the mean depth around the midpoint of the corresponding base.

As Fig.~\ref{mask123} shows, after projecting the fused keypoints onto the image plane of each camera, we draw the inferred isosceles trapezoids onto the mask image successively, based on the depth values. This hierarchical structure not only reflects the occlusion relationships between key parts, but also ensures that each point cloud of the human body belongs to at most one keypart. By traversing the depth image and the generated mask simultaneously, the point clouds corresponding to each keypart can be extracted, thereby facilitating the estimation of their position and orientation.

\subsection{Methodology Overview}

\begin{algorithm}[h]
%	\caption{Hierarchically Connected Sensing}
	\caption{Multi-View Active Sensing in HRI using Hierarchically Connected Tree}
	\label{algo}
	\KwIn{Color images $\bm{C}$ and depth images $\bm{D}$}
	\KwOut{human state $\bm{\chi}$}
	
	\For{each camera $i$}{
		$\bm{p}_{i,[t_n]},\bm{c}_{i,[t_n]} \leftarrow HRNet\left(\bm{C}_{i,[t_n]}\right)$ \\
		$\bm{d}_{i,[t_n]} \leftarrow \bm{D}_i\left(\bm{p}_{i,[t_n]}\right)$ \hfill $\rhd$ Keypoint 3D position \\
		$\mathcal{K}_i \leftarrow E\left(\bm{c}_{i,[t_{n-M}:t_n]}\right)$ \hfill $\rhd$ Keypoint presence \\
		$\mathcal{K}_j \leftarrow E\left(\bm{c}_{i,[t_{n-M}:t_n]}\right)$ \hfill $\rhd$ Keypart presence
	}
	
	\For{keypoint $k \in \bigcup \mathcal{K}_i$}{
		\For{camera $i$ that $k \in \mathcal{K}_i$}{
			$\bar{\bm{p}}^{\mathcal{C}_i}_{[t_n]} = d_{i,[t_n]} \mathbf{K}_i^{-1} \hat{\bm{p}}_{i,[t_n]}$ \\
		}
		$\bar{\bm{p}}^*_{k,[t_n]} \leftarrow \text{FuseP(}\bar{\bm{p}}_{[t_n]}\text{)}$ \hfill $\rhd$ Position fusion \\
		$c^*_{k,[t_n]} \leftarrow \text{UpdateC(}\bm{c}_{[t_n]}\text{)}$ \hfill $\rhd$ Confidence update \\
	}
	
	\For{each camera $i$}{
		$\bm{C}_i.\text{drawMask()}$ \\
		\For{point $\bar{pc} \in \bm{pc}_i$}{
			\If{$pc \in\bm{M}_i.\bm{pc}_j $}{
				$\bm{pc}_j.\text{push(}\bar{pc}\text{)}$ \hfill $\rhd$ Point cloud extraction
			}
		}
	}
	$\mathcal{T}.\text{init()}$ \hfill $\rhd$ Hierarchical tree initialization \\
	$\mathcal{T}.\text{augment()}$ \hfill $\rhd$ Connectivity supplement \\
	
	\For{keypart $j \in \mathcal{T}$}{
		$\bm{\chi}_j.\text{init()}$ \\
		$\text{ICP(}\bm{pc}_j,\bm{cy}_j\text{)}$ \hfill $\rhd$ Model registration \\
		$\bm{\chi}_j.\text{update()}$ \hfill $\rhd$ Constraint maintenance
	}
	
\end{algorithm}

Based on a hierarchically connected tree-structured model, our method framework is illustrated in Algorithm~\ref{algo}. This algorithm senses the human state $\bm{\chi}$ from color images $\bm{C}$ and depth images $\bm{D}$ captured by MCAV. State estimation is performed on each node of the tree structure in a hierarchical order, and subsequent information fusion is conducted considering anatomical constraints.

$\textit{Line 1-6}:$ For each camera $i$, HRNet is employed to infer the keypoint locations $\bm{p}_{i,[t_n]}$ and confidences $\bm{c}_{i,[t_n]}$ from color image $\bm{C}_i$. In conjunction with depth image $\bm{D}_i$, the inference is extended from 2D to 3D. And the set of reliabe keypoints $\mathcal{K}_i$ and keyparts $\mathcal{K}_j$ can be determined from the sliding window of scores $\bm{c}_{i,[t_{n-M}:t_n]}$.

$\textit{Line 7-21}:$ After aligning coordinates and timestamps, existing keypoint states from each camera are fused to enhance precision. The fused keypoints are projected onto the camera imaging plane, generating a more accurate human body mask $\bm{M}_i$. This mask is then applied to the depth image $\bm{D}_i$ to extract point clouds of keyparts $\bm{pc}_j$. Notably, the preprocessing steps, including voxel filtering, pass-through filtering, and Euclidean clustering, are applied to refine point clouds. And point clouds of the robotic arm are elimated using cylindrical envelope.

$\textit{Line 22-28}:$ The tree model $\mathcal{T}$ is structured based on $\bigcup \mathcal{K}_i$ and $\bigcup \mathcal{K}_j$, and is supplemented in order to maintain connectivity. Following the hierarchical order, the model cylinders $\bm{cy}_j$ and the data clouds $\bm{pc}_j$ are registered after initialization. The registration states $\bm{\chi}_j$ are updated based on biological constraints.

%%%%%%%%%%%%%%%%%%%%%%%%%%%%%%%%%%%%%%%%%%%%%%%%%%%%%%%%%%%%%%%%%%%%%%
\section{Experiments}
\label{section5}

\subsection{Experimental Setup}

To evaluate the enhancement brought by MCAV in human sensing, comparisons are made with other visual systems, including a single fixed camera, a single active camera and multiple fixed cameras. The robotic arm execute pick-and-place tasks with repetitive motions, which is common in industrial production. Three typical production scenarios are designed to validate the generality and applicability of our method. In the first scenario, a human operator performs simple assembly tasks within the robotic arm's workspace. The second scenario involves intensive human-robot interaction during assembly tasks, such as reaching out or placing objects. And in the third scenario, the human operator enters and exits the robotic arm's workspace multiple times. The comparison focuses on the keypart recognition and human pose estimation. Moreover, the performance of obstacle avoidance employing different visual systems is verified.

\begin{figure}[t]
	\centering
	%		\includesvg[width=1.0\textwidth]{Fig/fig1.svg}
	\includegraphics[width=1.0\columnwidth]{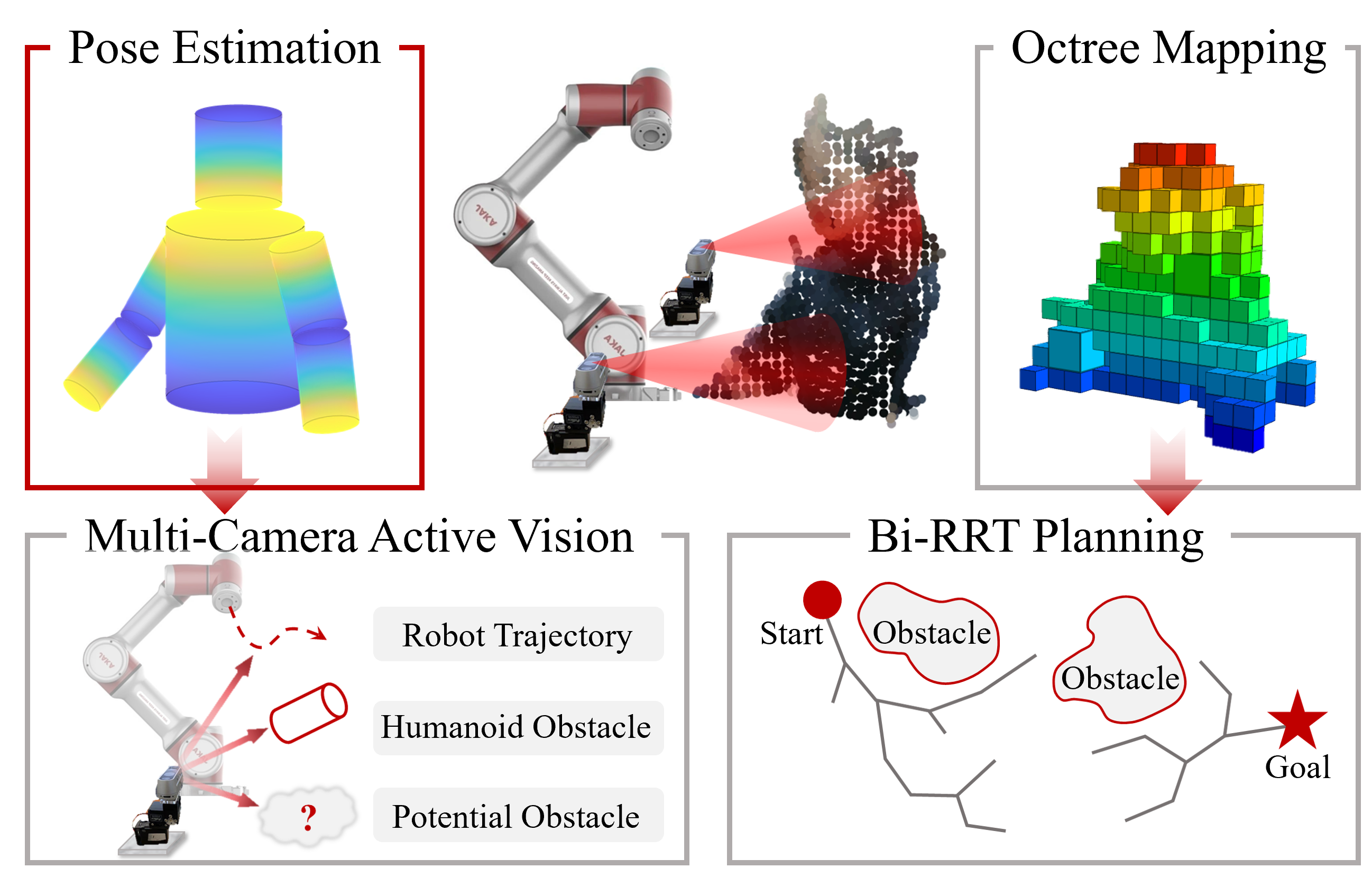}
	\caption{Experimental system module diagram}        
	\label{fig3}
\end{figure}

The robotic system primarily comprises four functional modules: human sensing, active vision scheduling, dynamic obstacle mapping and obstacle avoidance planning. In the human sensing module, our proposed method is employed, which is universally applicable to any number or configuration of cameras within the MCAV. As long as the real-time spatial poses of the cameras relative to the robotic arm are known, the environmental information acquired by each camera can be integrated to obtain a more comprehensive human sensing result.

In the MCAV, each camera is equipped with a rotation mechanism consisting of two vertically positioned, mutually orthogonal servos. This two-degree-of-freedom kinematic mechanism enables the camera to actively  rotate towards any direction within the spatial domain. To ensure the safety of HRI, the core of active vision strategy\cite{ref28} lies in orienting the cameras toward the direction of maximum collision probability. We formulate the scheduling problem as a Markov Decision Process (MDP) as follows:
\begin{equation}
	\tau=\arg _{\tau} \max \sum_{m=0}^{\text{M}} \gamma^m \ln \left(1-\hat{p}_\text{collide}^{m|m+1}\right)
\end{equation}
where $\tau:[0, t] \mapsto \mathbb{S}^2$ is the trajectory of the MCAV, $\gamma \in \left(0,1\right)$ is a decay coefficient. $p_\text{collide}^{m|m+1}$ represents the probability of collision between the robotic arm and the humanoid obstacle in the time interval $\left[t_m, t_{m+1}\right]$, and $\hat{p}_\text{collide}^{m|m+1}$ is its corresponding estimate. Each keypart is modeled as a cylindrical envelope characterized by its position, orientation, and the associated uncertainty. Given the desired trajectory of the robotic arm, the collision probability of each region can be determined. Employing the active vision strategy, the vision system can focus on crucial regions.

With the goal of avoiding human obstacles, we employ the spatial information to construct a real-time grid map. An octree data structure is chosen for mapping due to its efficient storage and retrieval of spatial data. Each link of the robotic arm is enveloped with a cylinder, and the FCL library is utilized for collision detection.  Upon detecting a potential collision between the robotic arm and the octree map, the trajectory of the robotic arm is dynamically replanned for obstacle avoidance. To strike a balance between solvability and efficiency, the Bi-RRT algorithm is employed for path planning. Subsequently, the planned path is pruned for smoothness and executed by the built-in actuator of the robotic arm.

The robotic arm employed in the experiment is the JAKA® Zu7 manipulator. The MCAV consists of two Intel® RealSense™ D435 cameras, each manipulated by two FEETECH® servos, allowing for focus adjustments from various perspectives. Our approach  is implemented within the ROS Noetic environment, and all processings are done in real time on a laptop featuring Intel® Core™ i7-12700H and NVIDIA® GEFORCE RTX™ 3060.

\subsection{Body Part Recognition}

\begin{figure}[t]
	\centering
	\includegraphics[width=0.5\textwidth]{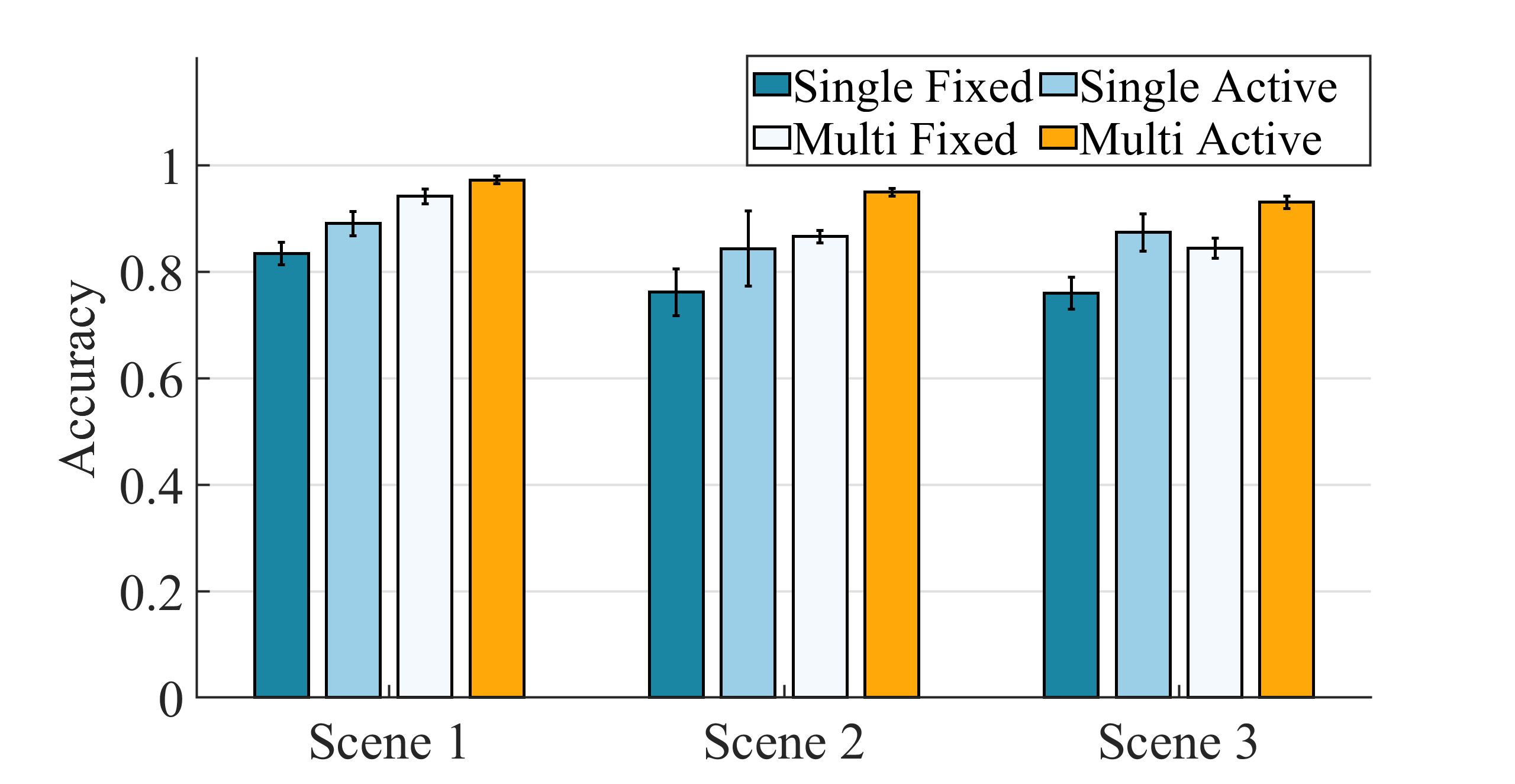}
	\caption{The mean and standard deviation of keypart recognition accuracy, computed from 10 trials.}        
	\label{accuracy}
\end{figure}

Determining the presence of human body parts can be regarded as a binary classification problem. Therefore, we use accuracy as evaluation measure to assess the performance of different visual systems, where $\text{Accuracy}=\frac{\text{True Positives}+\text{True Negatives}}{\text{Total Samples}}$. The ground truth of the samples is manually annotated from the recorded data of color images, and statistical results are obtained for each visual system in each scenario, see Fig.~\ref{accuracy} and Tab.~\ref{tab2}. The duration of each trial recording exceeds one minute, with the calibration images set to operate at a rate of 10 frames per second.

%	Determining the presence of human body parts can be regarded as a binary classification problem. Therefore, we use accuracy and recall as evaluation measure to assess the performance of different visual systems, where $\text{Accuracy}=\frac{\text{True Positives}+\text{True Negatives}}{\text{Total Samples}}$ and $\text{Recall}=\frac{\text{True Positives}}{\text{True Positives} + \text{False Negatives}}$, see Fig.~\ref{accuracy} and Table.~\ref{tab2}.

The average accuracy achieved by employing multi-camera active-vision system outperforms that of a single fixed camera, a single active camera, multiple fixed cameras by $31.69\%$, $19.84\%$ and $8.71\%$, respectively. In $\textit{Scene 1}$, where the human operator is centrally positioned in front of the robotic arm, less pronounced. However, the utilization of multiple cameras provides complementary FOV, resulting in accuracy exceeding $95\%$, compared to approximately $80\%$ with only one single camera. In $\textit{Scene 2}$, the occlusion caused by HRI underscores the effectiveness of active vision. It enables the visual system to focus on crucial areas and avoid occluded regions. The results indicate that using multiple cameras outperforms a single camera, and employing active vision surpasses fixed cameras. In $\textit{Scene 3}$, the entry and exit of the human operator, coupled with a large range of motion often exceeding the perception field of the visual system. While the accuracy with MCAV remains close to $90\%$, a single fixed camera achieves an accuracy of less than $60\%$.

\begin{table}[t]
	\caption{Average accuracy\label{tab2}}
	\centering
	\resizebox{1.0\columnwidth}{!}{
		\begin{tabular}{lcccc}
			\toprule
			& Scene 1    & Scene 2    & Scene 3    & Total                    \\ \midrule
			Multi Active   & $0.9721$  & $0.9398$  & $0.9300$  & $0.9473$              \\
			Multi Fixed    & $0.9413$  & $0.8749$  & $0.8477$  & $0.8880$              \\
			Single Active & $0.8901$  & $0.8651$  & $0.8738$  & $0.8763$              \\
			Single Fixed  & $0.8339$  & $0.8024$  & $0.7583$  & $0.7982$              \\ \bottomrule
		\end{tabular}
	}
\end{table}

%\begin{figure}[t]
%	\centering
%	\includegraphics[width=0.5\textwidth]{Fig/recall}
%	\caption{The mean and standard deviation of keypart recognition recall, computed from 10 trials.}        
%	\label{recall}
%\end{figure}
%
%\begin{table}[t]
%	\caption{Average recall\label{tab3}}
%	\centering
%	\resizebox{1.0\columnwidth}{!}{
%			\begin{tabular}{lcccc}
%					\toprule
%					& Scene 1    & Scene 2    & Scene 3    & Total          \\ \midrule
%					Multi Active    & $0.9529$  & $0.8960$  & $0.8542$  & $0.9010$             \\
%					Multi Fixed     & $0.9675$  & $0.8400$  & $0.7703$  & $0.8592$             \\
%					Single Active  & $0.8167$  & $0.7234$  & $0.6824$  & $0.7408$             \\
%					Single Fixed   & $0.7775$  & $0.7076$  & $0.5938$  & $0.6929$             \\ \bottomrule
%				\end{tabular}
%		}
%\end{table}

\subsection{Human Pose Estimation}

\begin{figure*}[htbp]
	\centering
	\subfloat[Scene 1]{
		\includegraphics[width=0.32\textwidth]{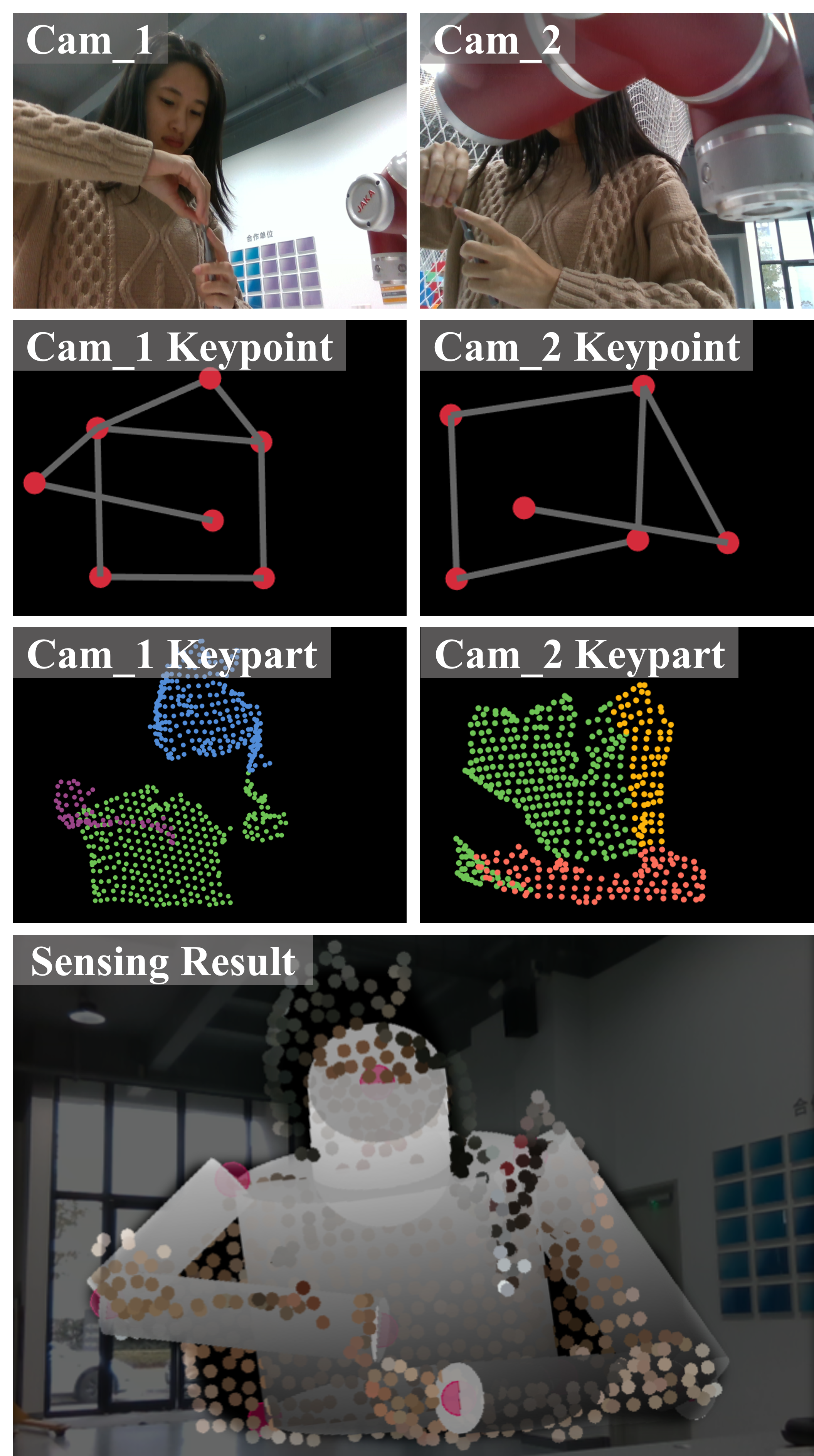}
	 	\label{result1}}
	\subfloat[Scene 2]{
		\includegraphics[width=0.32\textwidth]{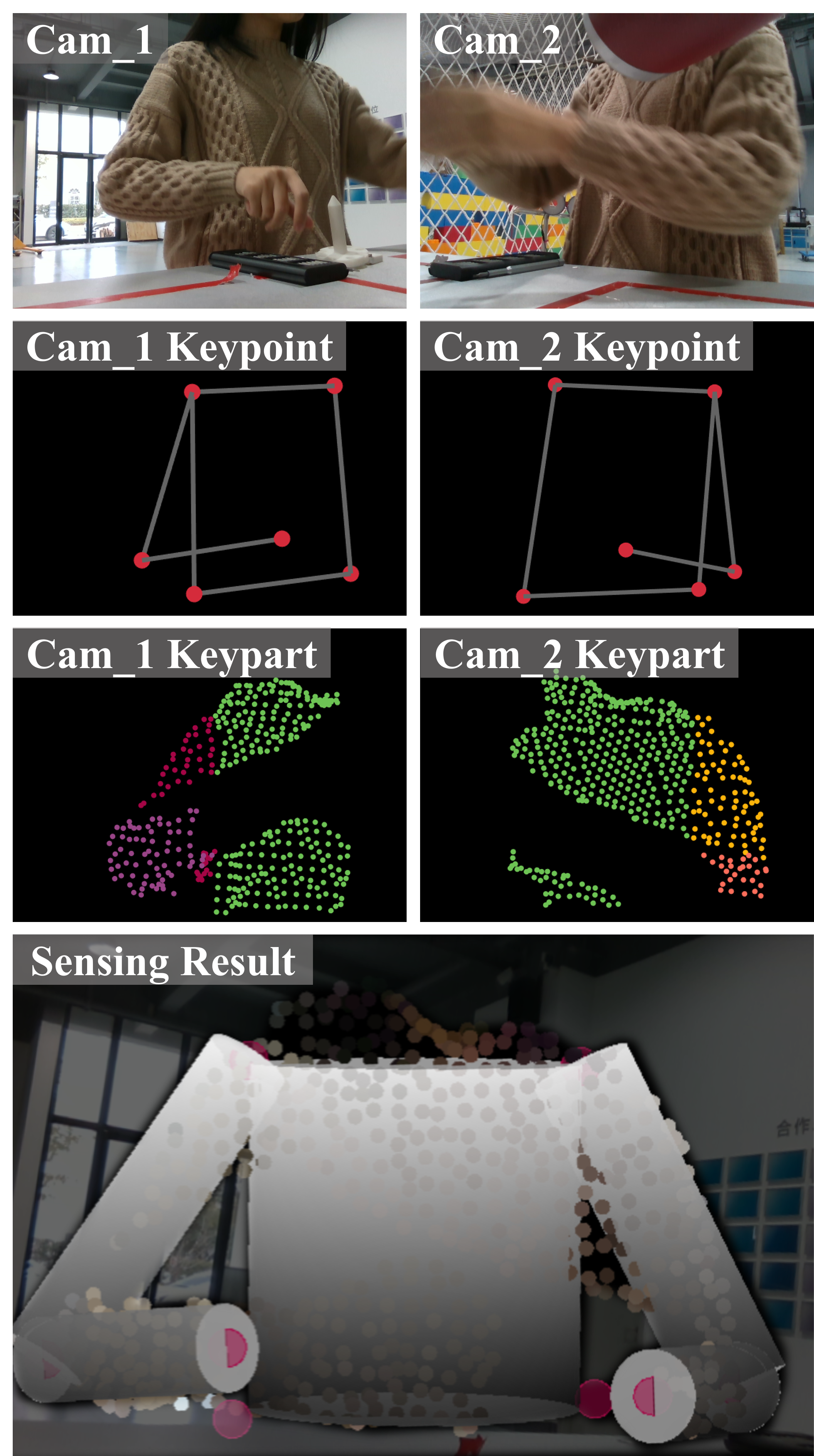}
		\label{result2}}
	\subfloat[Scene 3]{
		\includegraphics[width=0.32\textwidth]{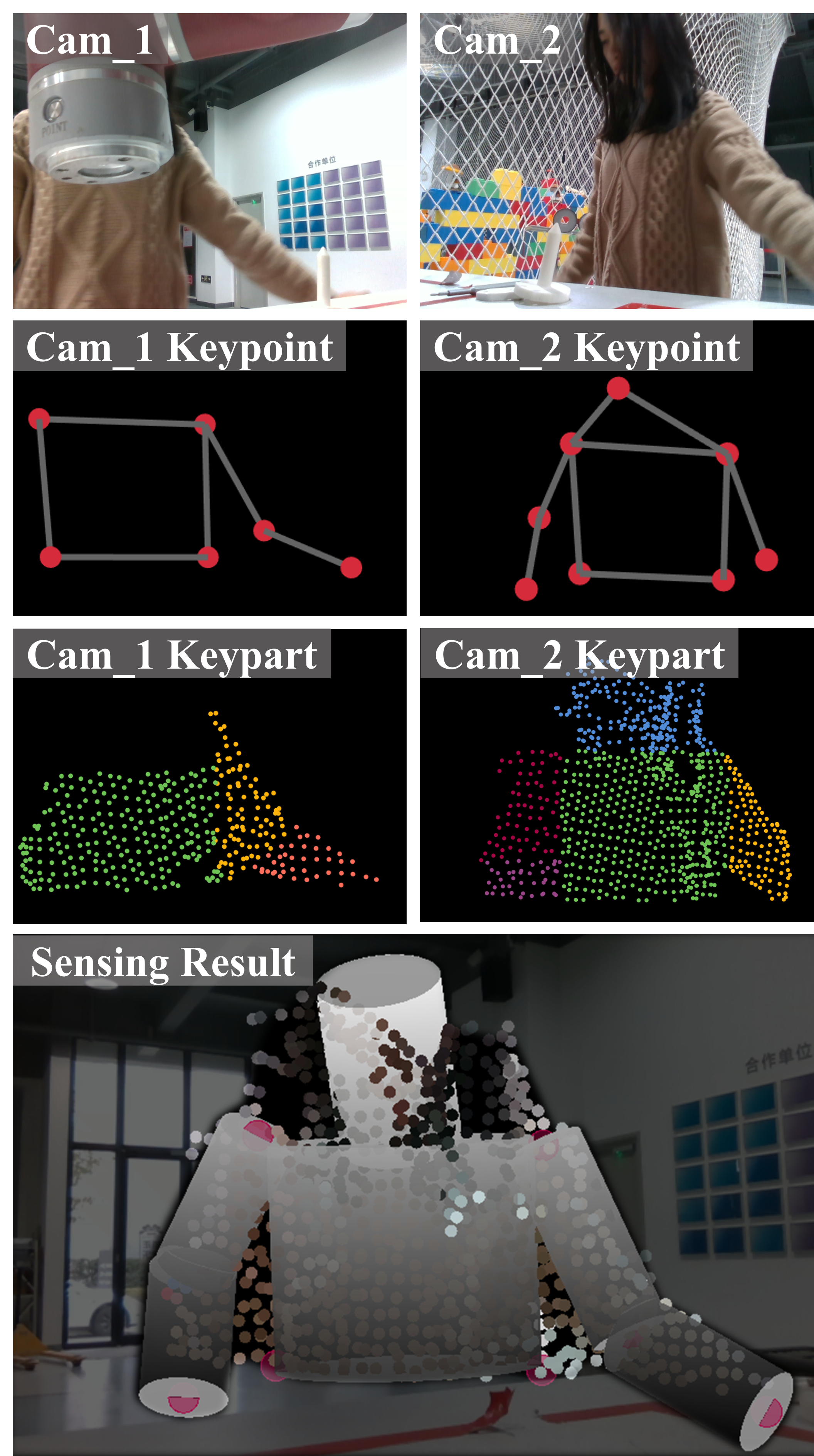}
		\label{result3}}
	\caption{Human pose estimation in practical HRI scenarios. The RGB images, keypoint 3D localization, keypart point cloud extraction, and the final cylindrical representation are illustrated in each subfigure. The human operator's actions include: during the execution of assembly tasks, obstructing the operation of the robotic arm by reaching out, and about to leave the robotic arm’s workspace.}
	\label{result}
\end{figure*}

To demonstrate the effectiveness of our proposed method in integrating visual information from multiple dynamic cameras for comprehensive and precise human sensing, we present typical perceptual results in Fig.~\ref{result}. The visual data captured by each camera is inherently constrained, with instances of localized FOV, self-occlusion or external occlusion. Using HRNet, the presence of each keypart and keypoint is determined, enabling the establishment of the tree topology. For example, in Fig.~\ref{result1}, self-occlusion of the left arm in $\textit{Cam 1}$’s FOV, the right arm being outside $\textit{Cam 2}$’s FOV, and occlusion of the head by robotic arm in $\textit{Cam 2}$’s FOV can be correctly identified. After localizing the keypoints, point clouds corresponding to the existing keyparts are extracted and combined. It can be inferred that the 3D localization of keypoints and the segmentation of keypart point clouds are consistent with actual scenarios. Finally, following the hierarchical order of the tree structure, the pose of the human body within the robotic arm’s workspace can be comprehensively sensed while maintaining anatomical constraints. As illustrated in Fig.~\ref{result}, existing body parts are perceived with precise estimation of their positions and orientations.

\subsection{Obstacle Avoidance}

To validate the efficacy of employing the hierarchically connected human sensing in the multi-camera active vision system for enhancing the safety of HRI, obstacle avoidance experiments are conducted on the robot arm performing pick-and-place tasks. The task execution cycle is divided in to four phases: pick-up ascent, pick-up descent, placement ascent, and placement descent. During each phase, a human operator dynamically extend their hand to obstruct the robot arm's task execution. Obstacle avoidance tests are performed for all four comparative visual systems, with each system undergoing three trials. The poses of human obstacles and the arm position obstructing task execution are consistent across the experiments. The success of obstacle avoidance and the trajectory of the robotic arm's end effector are recorded.

Table.~\ref{tab4} summarize the success of obstacle avoidance employing different visual systems. It's notable that employing multi-camera active vision achieves successful avoidance in every trial, while deploying one single camera only achieves a success rate of $16.67\%$. The utilization of multiple fixed cameras and one active camera, expanding the perceptual field and dynamically adjusting viewpoints of the camera, result in success rates of $50.00\%$ and $58.33\%$, respectively.

\begin{figure}[t]
	\centering
	\subfloat[Pick-up ascent]{
		\includegraphics[width=0.48\columnwidth]{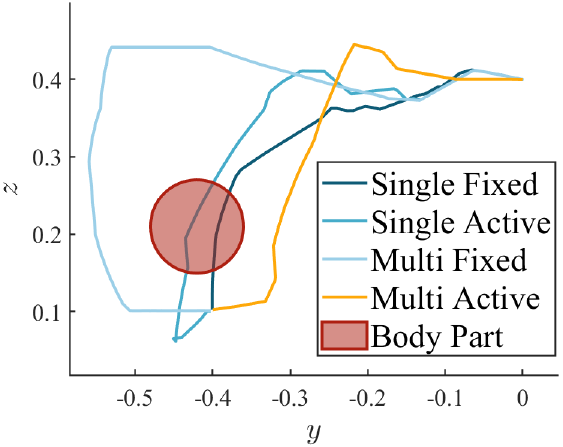}
		\label{avoid1}}
	\subfloat[Pick-up descent]{
		\includegraphics[width=0.48\columnwidth]{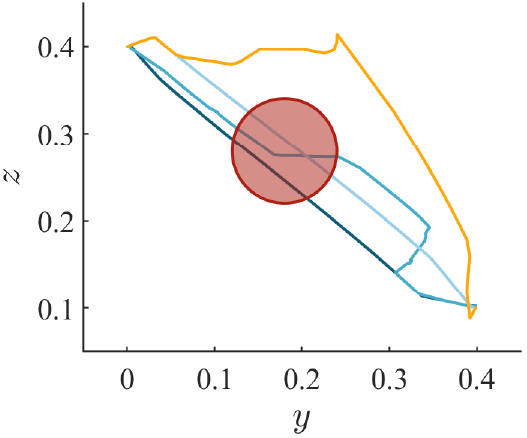}
		\label{avoid2}}\\
	\subfloat[Placement ascent]{
		\includegraphics[width=0.48\columnwidth]{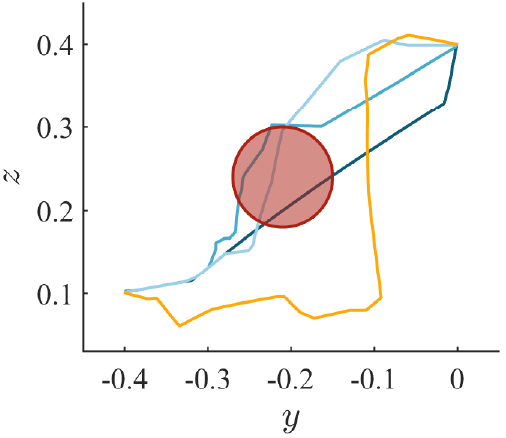}
		\label{avoid3}}
	\subfloat[Placement descent]{
		\includegraphics[width=0.48\columnwidth]{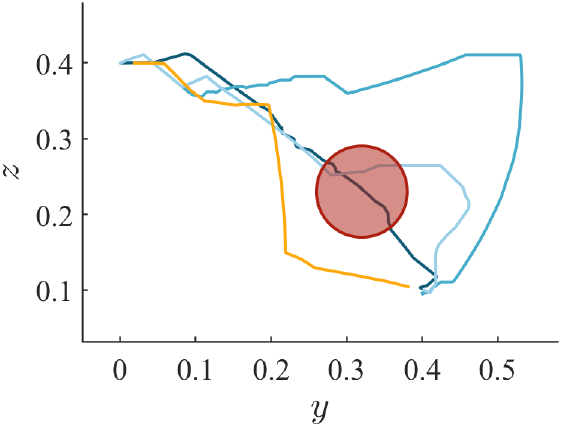}
		\label{avoid4}}\\
	\caption{Trajectory of robotic arm's end effector performing pick-and-place execution. (a)-(d) depict the trajectories of four comparative visual systems during corresponding phases. Each figure illustrates the projection of the trajectory onto the y-z plane of the base link coordinate system, and the projection of the cylindrical arm obstacle in this plane can be approximated as a circle.}
	\label{avoid}
\end{figure}

We select a representative cycle for each visual system during the execution of the pick-and-place task and plot the trajectory of the robotic arm's end effector in Fig.~\ref{avoid}. To depict a clearer obstacle avoidance process, the trajectory is projected onto the y-z plane of the base link coordinate system, with the projection of the cylindrical arm obstacle as a circle in this plane. Fig.~\ref{avoid1} and Fig.~\ref{avoid4} reveals that during the pick-up descent and placement ascent phases, only utilizing the multi-camera active vision can the robotic arm avoid the obstacles. The use of other visual systems, due to their limited perception, lead to collisions between the robotic arm and the arm obstacles. In the pick-up ascent and placement descent phases shown in Fig.~\ref{avoid2} and Fig.~\ref{avoid3}, besides the multi-camera active vision system, deploying multiple fixed cameras and a single active camera also achieve successful obstacle avoidance, respectively. However, their weaker human sensing capabilities constrain the solution space for obstacle avoidance planning, resulting in a larger pose variations of the robotic arm, and longer distances traveled by the end effecter to accomplish the avoidance goal. MCAV implementation not only achieve a 100\% success rate in avoiding arm obstacles, but also expands the perception and planning spaces, resulting in shorter paths during task execution.

\begin{table}[t]
	\caption{Obstacle Avoidance\label{tab4}}
	\centering
	\resizebox{1.0\columnwidth}{!}{
		\begin{tabular}{lccccc}
			\toprule
			& \makecell[c]{Pick \\ Ascent}    & \makecell[c]{Pick \\ Descent}    & \makecell[c]{Place \\ Ascent}    & \makecell[c]{Place \\ Descent}    & Total          \\ \midrule
			Muti Active    & $3$  & $3$  & $3$  & $3$ & $12$             \\
			Muti Fixed     & $2$  & $1$  & $1$  & $2$ & $6$             \\
			Single Active  & $1$  & $2$  & $2$  & $2$ & $7$             \\
			Single Fixed   & $1$  & $0$  & $0$  & $1$ & $2$             \\ \bottomrule
		\end{tabular}
	}
\end{table}

%%%%%%%%%%%%%%%%%%%%%%%%%%%%%%%%%%%%%%%%%%%%%%%%%%%%%%%%%%%%%%%%%%%%%%

\section{Conclusion}
\label{section6}

In the industrial manufacturing process involving robotic arms, human sensing is essential for ensuring the safety of human-robot interaction. To achieve comprehensive human perception, we propose a multi-view active sensing system. A hierarchically connected tree of human body is structured to integrate the visual information from each dynamic source. Compared to a single static camera, MCAV is more resilient to partial view and occlusion. Our method enhances keypart recognition accuracy from 79.82\% to 94.73\%, compared to employing a single static camera. Furthermore, our connected tree structure maintains the anatomical constraints while facilitating hierarchical pose estimation. Experimental results demonstrate that this leads to improved alignment between the captured point clouds and the cylindrical model. Additionally,  with the implementation of MCAV, the robotic arm achieves a 100\% success rate in obstacle avoidance. Dynamic human obstacles in different scenarios are perceived and avoided, significantly improving the safety of HRI. In the future, we plan to introduce velocity estimation and predictive positioning into the human model representation, achieving more detailed and effective human sensing.

%%%%%%%%%%%%%%%%%%%%%%%%%%%%%%%%%%%%%%%%%%%%%%%%%%%%%%%%%%%%%%%%%%%%%%

%%%%%%%%%%%%%%%%%%%%%%%%%%%%%%%%%%%%%%%%%%%%%%%%%%%%%%%%%%%%%%%%%%%%%%

\end{document}